\pdfoutput=1

\documentclass[11pt]{article}

\usepackage[final]{acl}

\usepackage{times}
\usepackage{latexsym}

\usepackage[T1]{fontenc}

\usepackage[utf8]{inputenc}

\usepackage{microtype}

\usepackage{inconsolata}

\usepackage{graphicx}
\usepackage{hyperref}
\usepackage{url}
\usepackage{microtype}
\usepackage{times}
\usepackage{latexsym}
\usepackage{soul}
\usepackage{amsmath}
\usepackage{array, booktabs}
\usepackage{bm}
\usepackage{cleveref}
\usepackage{soul}
\usepackage{multirow}
\usepackage{xpatch}
\usepackage{blindtext}
\usepackage{fdsymbol}
\usepackage{microtype}
\usepackage{hyperref}
\usepackage[export]{adjustbox}
\usepackage{textcomp}
\usepackage{xparse}
\usepackage{enumitem}
\usepackage{makecell}
\usepackage{subcaption}
\usepackage{colortbl}
\usepackage{tcolorbox}
\usepackage{xparse}
\usepackage{stfloats}
\usepackage{arydshln}
\usepackage{bbm}

\newcommand{\specialcell}[2][c]{%
  \begin{tabular}[#1]{@{}c@{}}#2\end{tabular}}
  
\usepackage{tabularx}

\usepackage{colortbl}

\usepackage{xcolor,pifont}
\newcommand*\colourcheck[1]{%
  \expandafter\newcommand\csname #1check\endcsname{\textcolor{#1}{\ding{52}}}%
}
\colourcheck{blue}
\colourcheck{green}
\colourcheck{red}

\usepackage[colorinlistoftodos]{todonotes}
\newcommand{\xmark}{\textcolor{red}{\ding{55}}}%

\usepackage{listings}

\definecolor{keycolor}{rgb}{0,0,0.8}    
\definecolor{stringcolor}{rgb}{0.5,0,0} 
\definecolor{numbercolor}{rgb}{0.5,0,0.5} 

\definecolor{verylightgray}{rgb}{0.9,0.9,0.9}

\lstdefinelanguage{json}{
    basicstyle=\normalfont\ttfamily,
    showstringspaces=false,
    breaklines=true,
    frame=lines,
    backgroundcolor=\color{verylightgray},
    literate=
     *{0}{{{\color{numbercolor}0}}}{1}
      {1}{{{\color{numbercolor}1}}}{1}
      {2}{{{\color{numbercolor}2}}}{1}
      {3}{{{\color{numbercolor}3}}}{1}
      {4}{{{\color{numbercolor}4}}}{1}
      {5}{{{\color{numbercolor}5}}}{1}
      {6}{{{\color{numbercolor}6}}}{1}
      {7}{{{\color{numbercolor}7}}}{1}
      {8}{{{\color{numbercolor}8}}}{1}
      {9}{{{\color{numbercolor}9}}}{1}
      {:}{{{\color{keycolor}{:}}}}{1}
      {,}{{{\color{keycolor}{,}}}}{1}
      {\{}{{{\color{keycolor}{\{}}}}{1}
      {\}}{{{\color{keycolor}{\}}}}}{1}
      {[}{{{\color{keycolor}{[}}}}{1}
      {]}{{{\color{keycolor}{]}}}}{1}
      {"}{{{\color{stringcolor}{"}}}}{1},
}

\definecolor{lightgray}{rgb}{0.94,0.95,0.95}
\lstdefinelanguage{json2}{
    basicstyle=\normalfont\fontfamily{pcr}\selectfont,
    showstringspaces=false,
    breaklines=true,
    frame=lines,
    backgroundcolor=\color{lightgray},
    literate=
     *{0}{{{\color{numbercolor}0}}}{1}
      {1}{{{\color{numbercolor}1}}}{1}
      {2}{{{\color{numbercolor}2}}}{1}
      {3}{{{\color{numbercolor}3}}}{1}
      {4}{{{\color{numbercolor}4}}}{1}
      {5}{{{\color{numbercolor}5}}}{1}
      {6}{{{\color{numbercolor}6}}}{1}
      {7}{{{\color{numbercolor}7}}}{1}
      {8}{{{\color{numbercolor}8}}}{1}
      {9}{{{\color{numbercolor}9}}}{1}
      {:}{{{\color{keycolor}{:}}}}{1}
      {,}{{{\color{keycolor}{,}}}}{1}
      {\{}{{{\color{keycolor}{\{}}}}{1}
      {\}}{{{\color{keycolor}{\}}}}}{1}
      {[}{{{\color{keycolor}{[}}}}{1}
      {]}{{{\color{keycolor}{]}}}}{1}
      {"}{{{\color{stringcolor}{"}}}}{1},
}

\definecolor{mygray}{rgb}{0.95, 0.95, 0.95}
\definecolor{myblue}{rgb}{0.41, 0.50, 0.57}
\definecolor{greyblue}{RGB}{177,221,240}
\definecolor{lightgold}{RGB}{249,247,237}
\definecolor{peacockblue}{RGB}{27,161,226}

\tcbuselibrary{listings,skins,breakable}

\tcbset{
    enhanced,
    breakable,
    colback=mygray, 
    colframe=myblue, 
    boxrule=0.2mm, 
    arc=1mm, 
    top=10pt,
    bottom=10pt,
    left=10pt,
    right=10pt,
    listing only,
    listing options={
        basicstyle=\ttfamily\small, 
        language=Java, 
    },
}

\newcommand*{\affaddr}[1]{#1} 

\newcommand{\method}[1]{\textsc{Lam Simulator}}

\crefformat{section}{\S#2#1#3} 
\crefformat{subsection}{\S#2#1#3}
\crefformat{subsubsection}{\S#2#1#3}

\definecolor{gold}{rgb}{0.83, 0.69, 0.22}
\iftrue

\NewDocumentCommand{\steeve}
{ mO{} }{\textcolor{gold}{\textsuperscript{\textit{Steeve}}\textsf{\textbf{\small[#1]}}}}

\else
\newcommand{\steeve}[1]{}

\fi

\title{\method~: Advancing Data Generation for Large Action Model Training via Online Exploration and Trajectory Feedback}

\author{
\text{Thai~Hoang},
\text{Kung-Hsiang~Huang \thanks{\, Core Contributors.}},
\textbf{Shirley~Kokane \footnotemark[1]},
\text{Jianguo~Zhang},
\text{Zuxin~Liu},
\text{Ming~Zhu},
\\
\textbf{Jake~Grigsby}, 
\textbf{Tian~Lan}, 
\textbf{Michael S Ryoo},
\textbf{Chien-Sheng Wu},
\textbf{Shelby Heinecke},
\textbf{Huan~Wang},
\\
\textbf{Silvio~Savarese},
\textbf{Caiming Xiong},
\textbf{Juan Carlos Niebles}
\\
\affaddr{Salesforce AI Research, USA} \\
\affaddr{\{thai.hoang, kh.huang, skokane\}@salesforce.com}
}

\begin{document}
\maketitle

\begin{abstract}
Large Action Models (LAMs) for AI Agents offer incredible potential but face challenges due to the need for high-quality training data, especially for multi-steps tasks that involve planning, executing tool calls, and responding to feedback. To address these issues, we present \method~, a comprehensive framework designed for online exploration of agentic tasks with high-quality feedback. Our framework features a dynamic task query generator, an extensive collection of tools, and an interactive environment where Large Language Model (LLM) Agents can call tools and receive real-time feedback. This setup enables LLM Agents to explore and solve tasks autonomously, facilitating the discovery of multiple approaches to tackle any given task. The resulting action trajectory data are then used to create high-quality training datasets for LAMs. Our experiments on popular agentic benchmarks, ToolBench and CRMArena, highlight the effectiveness of \method~: models trained with self-generated datasets using our framework achieve significant performance gains, up to a 49.3\% improvement over their original baselines. \method~ requires minimal human input during dataset creation, highlighting \method~'s efficiency and effectiveness in speeding up development of AI agents.

\end{abstract}

\section{Introduction}
Large Action Models (LAMs)~\citep{zhang2024xlam,xu2024lemur,liu2024apigen} are an advanced type of Large Language Model, specifically optimized for tool usage, reasoning, and function calling. Recent advancements have propelled their capabilities, making them integral to applications such as AI agents and task automation. LAMs benefit from specialized training for enhanced performance in agent applications. As use cases grow, the demand for more accurate models will continue to increase.

Current approaches for creating LAMs include prompt engineering, incorporating additional contextual information into prompts, Supervised Fine-Tuning (SFT), Reinforcement Learning from Human Feedback (RLHF) \cite{ouyang2022training}, among others. Most of these methods, however, rely heavily on manual data curation, a process that is both time-consuming and expensive.

\begin{figure*}[ht]
\centering
\includegraphics[width=\textwidth]{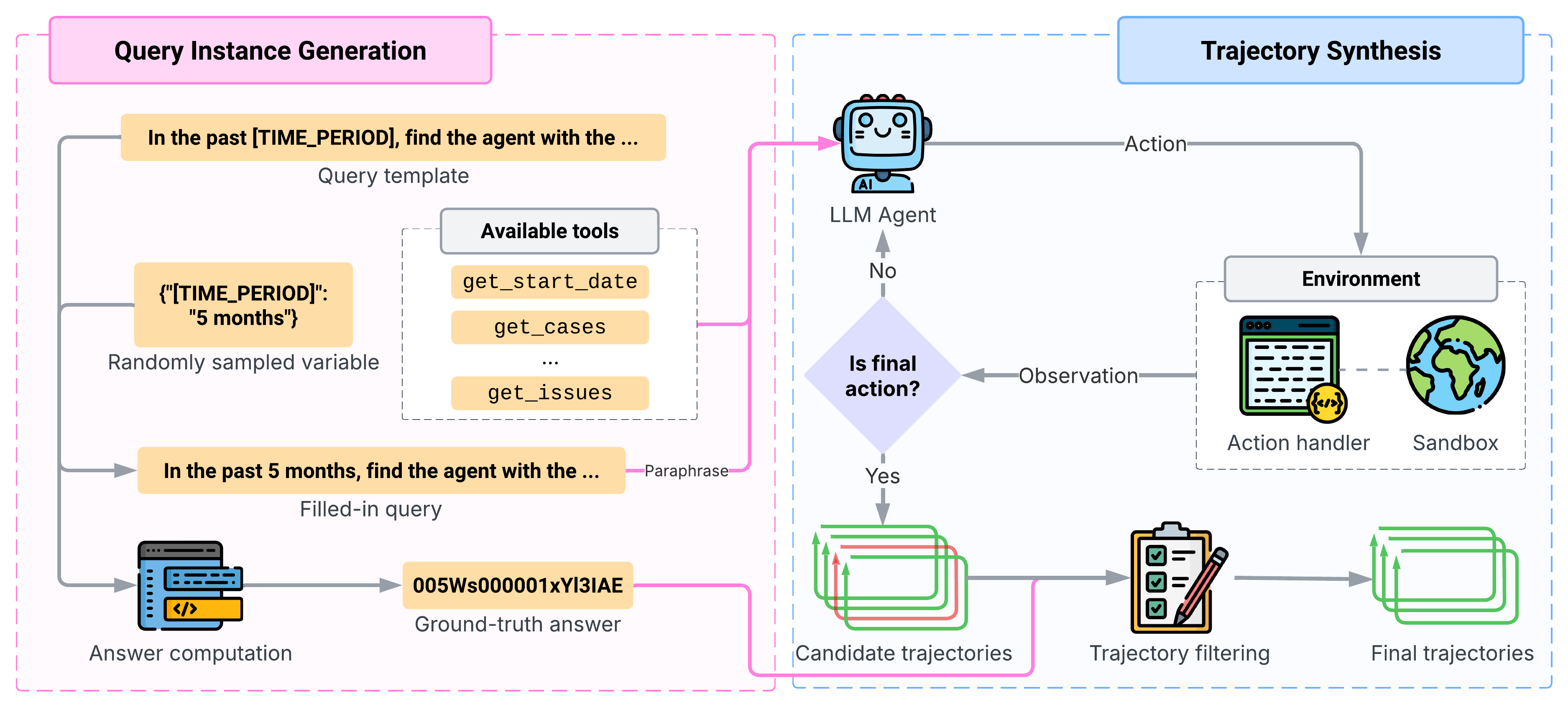}
\caption{\small Overview of the \method~, illustrating the framework's main components, their interactions, and emphasizing the its ability to generate tool-use data, execute functions, and evaluate results. \textbf{Query Instance Generation} is responsible for generating tasks, including creating user queries, preparing available tools for the LLM Agent, and dynamically computing ground-truth answers. \textbf{Trajectory Synthesis} manages the generation of agent trajectories by facilitating interactions between the LLM Agent and various environments, and by providing feedback and quality assessments for the generated trajectories.}
\label{fig:framework_overview}
\end{figure*}

To address this challenge, the use of LLM Agents to explore environments autonomously has emerged as a promising method to reduce the need for human labeling and annotation in agent model development. Recent studies, such as ToolTalk ~\citep{farn2023tooltalk}, WebArena ~\cite{zhou2023webarena}, APIGen ~\cite{liu2024apigen}, and Learn-by-Interact ~\cite{su2025learnbyinteractdatacentricframeworkselfadaptive}, have demonstrated the ability to automatically generate action trajectory data for agent learning and evaluation. However, ToolTalk is limited to specific tasks that are curated or filtered by humans. WebArena offers a very limited action space within Web domain. APIGen and Learn-by-Interact, while showing considerable potential in generating agentic data, are limited to the heavy usage of LLMs to assess data quality, thus introducing a notable amount of uncertainty, which is an important issue to consider.

Given these limitations, we introduce \method~, a comprehensive framework designed to enhance data generation for agent learning through exploration. As shown in Figure \ref{fig:framework_overview}, \method~ employs a template-filling strategy to dynamically create queries. Users only need to develop a series of query templates and descriptions for query parameters, along with logic to compute task answers using these parameters. A large language model (LLM) then creates values to fill into the query parameters in real-time and populates the templates to form real user queries. These generated queries are then given to LLM Agents, which explore solutions using a provided set of tools that might differ from those used to generate the ground-truth answers. As the agent tackles each task by making a series of function calls, \method~ provides immediate feedback. This enables a smooth interaction between the agent and its environment, allowing agents to freely explore problem-solving and rectify their actions through real-time feedback. Once the agent finishes its task, we apply thorough filtering to the generated trajectories, utilizing both the agent's data and pre-computed ground-truth answers. This process facilitates the creation of diverse action trajectory datasets suitable for LAM training.

Our testing on well-known agentic benchmarks, ToolBench ~\cite{qin2023toolllm} and CRMArena ~\cite{huang-etal-2025-crmarena}, demonstrates the high quality of data produced using \method~. When fine-tuning top-performing models with data they generated themselves, we observed a pass rate increase of 4.1\% for \texttt{gpt-4o} on ToolBench, and a 24.1\% increase on CRMArena. In addition, models with lower baseline performance showed significant improvement, with \texttt{mixtral-8x7b-inst} achieving over a 19.3\% increase on ToolBench, and \texttt{gpt-4o-mini} improving by 49.3\% on CRMArena. These results clearly demonstrate substantial performance enhancements resulting from exploration through \method~, underscoring the effectiveness of our framework across diverse environments. Our experiments demonstrate the remarkable effectiveness of the \method~ in improving model performance and identifying and addressing model weaknesses in an automated manner.

Our main contributions are:

\noindent\textbf{1.} We introduce a robust and high-quality data generation framework that features dynamic answer computation, enabling the creation of diverse, reliable agent trajectories for both single-turn and multi-turn scenarios.

\noindent\textbf{2.} We assemble an extensive library of over 3,000 interactive tools and design 36 comprehensive high-level tasks for ToolBench and CRMArena. This supports the generation of thousands of unique tasks. Our approach can be easily support new environments using the same methodology.

\noindent\textbf{3.} We conduct thorough benchmarking of the LAM Simulator on challenging, realistic agentic benchmarks. Our results show substantial performance gains over base models, and direct comparisons with existing baselines further highlight the significant improvements delivered by our framework.

\section{Related Work}
With the rapid evolution of Large Language Models (LLMs), there has been a significant increase in their application to tool-use and function-calling scenarios. Enhancing the capabilities of LLMs~\citep{achiam2023gpt,anthropic2024claude,dubey2024llama,zhang2024xlam} with external tools allows them to go beyond the limitations of their static parametric knowledge and text-based input-output interfaces. This extension enables them to access real-time information, leverage external reasoning systems, and perform meaningful actions in dynamic environments.

Recently, open-sourced research has focused increasingly on enhancing the efficiency of LLMs in tool-use contexts~\citep{qin2023toolllm,chen2023fireact,liu2024agentlite,zhang2024agentohana,bfcl,zhang2025actionstudio,ma2024tacolearningmultimodalaction}, while also exploring various prompting and training strategies to improve their performance in agentic tasks. Prominent prompting techniques like Chain of Thought (CoT)~\citep{wei2022chain}, Reflection~\citep{shinn2024reflexion}, and ReACT~\citep{yao2023react} have garnered attention. While initial efforts centered on In-Context Learning (ICL)—where pre-trained LLMs were prompted with API specifications and tool-use examples—current approaches are increasingly incorporating fine-tuning methods to enhance model accuracy.

Moreover, popular agent environments such as ToolEval ~\citep{qin2023toolllm}, AgentBench~\citep{liu2023agentbench}, WebArena~\citep{zhou2023webarena}, OSworld~\citep{xie2024osworld}, AgentBoard~\citep{ma2024agentboard}, SpecTool ~\cite{kokane2024spectoolbenchmarkcharacterizingerrors}, and $\tau$-bench~\citep{yao2024tau} facilitate agent interactions and evaluations within various scenarios such as web navigation, shopping, games, and computer environments. ToolEval offers a large test set with diverse scenarios, but the heavy reliance on LLMs for evaluation introduces undesired and uncontrollable variability. AgentBoard designs two multi-turn environments for tool query and operations, but they only contain 100 user queries in total and do not include real-time feedback. WebArena, OSWorld, and $\tau$-bench simulates conversations between a user and a language agent, providing API tools and policy guidelines with some limited domains: as web, OS, retail and airline.

The development of data generation pipelines for agentic learning has also seen significant advancements in recent years. ToolBench ~\citep{qin2023toolllm} introduced a wide range of tasks and tools but often suffers from lower data quality because it relies heavily on LLMs to create and evaluate tasks, which can introduce noise. APIGen ~\citep{liu2024apigen} improves data quality by using multi-layered evaluation schemes, but it is limited to handling only single-turn conversations. ETO ~\citep{song2024trialerrorexplorationbasedtrajectory} uses a fixed set of instructions and labeled data to generate trajectories, making it unsuitable for situations lacking pre-existing training data. Lastly, Learn-by-Interact ~\citep{su2025learnbyinteractdatacentricframeworkselfadaptive} brings diversity to tasks by constructing them backward from desired outcomes, but its dependence on comprehensive documentation limits its usefulness in domains with sparse references.

Our proposed framework, \method~, addresses these limitations and introduces several key innovations. Unlike previous approaches, \method~ natively supports complex multi-turn conversations, rather than being confined to single-turn interactions. It also allows an open-ended action space with limited context or documentation needed, meaning the agent can use any combination of tools without restrictions. We automate data generation at scale, combining a vast collection of tools with carefully-crafted, high-quality task templates. Furthermore, all evaluations in our framework are fully programmatic, ensuring the consistency and accuracy of the constructed datasets by avoiding the variability introduced by LLM-based evaluation. Table \ref{tab:framework_features} summarizes how our approach advances AI agent trained compared to previously leading systems.

\begin{table*}[ht]
\centering
\begin{tabularx}{\textwidth}{lXXXX}
\hline
\specialcell{Framework} & \specialcell{Multi-turn} & \specialcell{Open \\ Action} & \specialcell{Automated \\ Data Gen} & \specialcell{Program- \\ matic Evals} \\ \hline
ToolBench ~\cite{qin2023toolllm} & \greencheck & \greencheck & \greencheck & \xmark \\
APIGen ~\cite{liu2024apigen}& \xmark & \greencheck & \greencheck & \xmark \\
Learn-by-Interact ~\cite{su2025learnbyinteractdatacentricframeworkselfadaptive} & \greencheck & \greencheck & \greencheck & \xmark \\
\textbf{\method~ (ours)} & \greencheck & \greencheck & \greencheck & \greencheck \\ \hline
\end{tabularx}
\caption{\small \method~ compared to prior frameworks. \textbf{Multi-turn} indicates support for multi-turn settings, \textbf{Open Action} indicates if agent’s actions space are predefined or open, \textbf{Automated Data Gen} indicates automated training data generation capabilities, and \textbf{Programmatic Evals} indicates if ALL evaluators in the framework are using a programmatic approach without using LLMs.}
\label{tab:framework_features}
\end{table*}

\section{\method~}

We propose \method~ to enable LLM Agents to autonomously explore and enhance their problem-solving skills. We first construct query instances, each of which details the goals the LLM Agents should accomplish and the available tools (\Cref{subsec:query_gen}). Based on the given query instance, LLM Agents self-synthesize trajectories by iteratively interacting with the environment until the final state is reached (\Cref{subsec:traj_synthesis}). Finally, the generated trajectories are filtered based on ground-truth answers, and subsequently used for training LLM Agents. \Cref{fig:framework_overview} shows an overview of our framework. We describe how each component works in the following subsections. \looseness=-1

\subsection{Preliminaries}
The task defined by each query instance can be conceptualized as a Partially Observable Markov Decision Process (POMDP), defined by the tuple $(\mathcal{U}, \mathcal{S}, \mathcal{A}, \mathcal{O}, \mathcal{T})$. Here, $\mathcal{U}$ denotes the user query space, $\mathcal{S}$ represents the state space, $\mathcal{A}$ is the action space, $\mathcal{O}$ refers to the observation space, $\mathcal{T}: \mathcal{S} \times \mathcal{A} \to \mathcal{S}$ is the state transition function.

\subsection{Query Instance Generation}
\label{subsec:query_gen}
Query instances form the starting point for trajectory synthesis, comprising three elements: a user query $u \in \mathcal{U}$, a set of available tools $\mathcal{F}$
, and a ground-truth answer $y$. The user query and tools initiate agents' iterative self-exploration, while the ground-truth answer is used to verify the validity of the resulting trajectory, as detailed in \Cref{subsec:traj_synthesis}. Below, we illustrate the details of each element.

\paragraph{User Query Construction.} User queries are natural-language questions that specify the objectives the agent must achieve. We begin constructing user queries by manually creating query templates with placeholders. The variation in language style can be later managed through the use of LLMs for paraphrasing. Once the query templates are in place, we utilize LLMs to sample values for these placeholders. The queries that result from filling in the placeholders are paraphrased and finalized as user queries.\footnote{In this work, we employ \texttt{gpt-4o} to generate appropriate placeholder values and paraphrase texts.} A detailed example for a Query Instance is covered in \Cref{subsec:example_query_instance}.

\paragraph{Ground-truth Answer Computation.} In developing query templates, each template is linked to a sequence of tool usages aimed at deriving a ground-truth answer. This programmatic approach ensures the accurate production of answers. Here, the ground-truth-answer will later be used to compare with agent's answer during trajectory filtering step. Nonetheless, the tools used to generate the ground-truth answer are unlikely to align with those later made accessible to LLM Agents for exploration. This potential discrepancy is intentional, as we encourage agents to engage in exploratory problem-solving with all the available tools, rather than merely replicating a predefined sequence of tool calls. Consequently, a successful strategy may either align with these hidden solution paths or comprise an alternative series of actions that achieve the same objective.

\paragraph{Available Tools.}
The available tools for each user query are dependent on the environment. We describe the details of tools in \Cref{sesc:applications}.

\subsection{Trajectory Synthesis}
\label{subsec:traj_synthesis}

\paragraph{Iterative Self-exploration.}
Given a user query $u \in \mathcal{U}$, for each time step $t$ at state $s_t \in \mathcal{S}$, the agent selects an action $a_t = (f, p)\in \mathcal{A}$ by choosing an appropriate tool from the available set of tools $f \in \mathcal{F}$ and corresponding tool-call arguments $p$. This tool is used to interact with the environment, resulting in an observation $o_t \in \mathcal{O}$ after the function is executed. This process continues iteratively until a final state is reached. The final state is achieved under one of two conditions: (1) the agent performs an action that returns a result to the user, such as the \texttt{submit} function in CRMArena \cite{huang-etal-2025-crmarena}; or (2) the trajectory of actions exceeds the predefined maximum number of steps.

To guarantee the proper execution of tool calls and enable agents to learn from significant feedback from the environment, we introduced an action handler. This handler checks the structure, syntax, and validity of the tool call to ensure responses are in the correct required format, and also to avoid hallucinations like fabricated tool names or malformatted tool-call arguments. Second, it retrieves the error message from the sandbox and sends it back to the agent for correction, while also maintaining a record of the error history for the trajectory filtering stage.

\paragraph{Trajectory Filtering.}
Trajectory filtering guarantees that the resulting paths are both valid and useful for training. We accomplish this by using string matching to compare the final response $y'$ of each trajectory with the actual answer $y$. Any discrepancies ($y \neq y'$) result in exclusion from the selected set. Furthermore, to ensure the quality of our training dataset, we selectively include trajectories that meet the following criteria: (1) the LLM Agent completes the process without any errors, or (2) if any errors occur during tool usage, they are rectified in the subsequent action. For example, if the agent incorrectly applies a parameter to a tool at action $a_t$, we required that it is corrected in the next action, $a_{t+1}$.

\paragraph{Programmatic Evaluation.} Our framework includes the Action Handler for assessing actions and Trajectory Filtering for monitoring trajectory quality. This ensures the selection of trajectories that demonstrate effective tool use and accuracy, enhancing our confidence in the data quality. By pre-calculating the ground-truth answer and leveraging the components within Action Handlers and Trajectory Filtering, we can simulate high-quality feedback for each agent's action and trajectory, eliminating the inconsistencies and noisy evaluations often associated with using LLMs.

\subsection{Agent Training}
After the trajectory synthesis process, we train our LLM Agents on these self-generated trajectories. This approach offers two advantages over using only the gold-standard trajectories from \Cref{subsec:query_gen}. First, training on agent-explored trajectories, which may include errors and corrections in subsequent iterations, allows the model to explicitly learn error recovery strategies. This is crucial for real-world deployments where unexpected inputs or tool states are common. Second, our approach allows the model to encounter and adapt to a wider range of scenarios, including potential interactions with previously unseen tools or alternative, valid solution paths. As empirically demonstrated in \Cref{sec:exps}, exposing the model to this broader range of interactions during training notably enhances its generalization abilities and its robustness in utilizing tools and completing agent-based tasks.\looseness=-1

\subsection{Generalizability}
The proposed \method~ framework is easily generalizable to a wide range of agentic environments and tasks. To support a new set of tasks and environments, one only needs to design a set of query templates reflecting the specific task goals and establish the associated mappings from these templates to sequences of tool calls that yield the ground-truth answers. The capacity to rapidly adapt to novel tasks not only underscores the generalizability of \method~ but also highlights its potential as a universal tool for enhancing problem-solving capabilities of AI Agents.
\section{Applications}
\label{sesc:applications}
\method~ is engineered to be seamlessly adaptable to a wide range of scenarios, showcasing its potential in advancing contemporary models. To demonstrate the effectiveness of \method~, we supported its application to two prominent environments for agentic tasks: ToolBench \cite{qin2023toolllm} and CRMArena \cite{huang-etal-2025-crmarena}. ToolBench emphasizes generic tool-use capability, while CRMArena delves into complex and realistic Customer Relationship Management (CRM) scenarios. Our goal is to exhibit LLM Agent’s self-improvement capability via explorative processes through \method~.

\subsection{ToolBench}
\paragraph{Tool collection creation.}
\label{subsubsec:toolbench_tools}
We leveraged ToolBench~\cite{qin2023toolllm}'s extensive repository, which encompasses 16,464 REST APIs sourced from the RapidAPI Hub. Although this collection is notably extensive, we encountered numerous entries that were either non-functional or inadequately documented. To enhance the quality and reliability of our exploration, we conducted a thorough clean-up process as detailed in \Cref{subsec:cleanup_toolbench_tools}. This refinement led us to a more manageable and useful collection comprising 3,420 effective tools. Additionally, given that the ToolBench's tools collection necessitated the use of tools from various providers, availability issues could arise during exploration. To address these challenges and ensure stability in exploration, our team constructed a suite of 57 tools. These tools encompass critical domains, such as Data, Science, Entertainment, and Tool Usage. This strategic approach aims to reduce reliance on external providers, thereby enhancing the reliability and consistency of tool availability during exploration.

\paragraph{Query Instances.}
\label{subsubsec:toolbench_query_instances}
We devised 30 query templates based on instances from the ToolBench training dataset, each addressing objectives such as retrieving movie details or housing property searches. For these tasks, we developed sequences of tool calls to generate solutions utilizing our in-house tools, as detailed above. This process yielded 400 unique query instances, with each instance consists of a paraphrased fill-in query with parameters produced by LLMs, a pre-determined ground-truth answer, and a set of tools for exploration. These tools include either those used to compute the solutions or alternative options, along with supplementary tools intended to challenge the decision-making capabilities of agents.

\subsection{CRMArena}
\paragraph{Tool collection creation.}
To facilitate exploration in solving the tasks for CRMArena, relevant tools necessary for the four supported tasks were extracted, representing 15 out of the 25 tools available in CRMArena~\cite{huang-etal-2025-crmarena}. The remaining 10 tools were deliberately left unseen to rigorously test the system's adaptability in out-of-domain scenarios.

\paragraph{Query Instances.}
\label{subsec:crmarena_tasks}
From the six finely crafted tasks derived from their framework: New Case Routing (NCR), Handle Time Understanding (HTU), Monthly Trend Analysis (MTA), Best Region Identification (BRI), Transfer Count Understanding (TCU), and Top Issue Identification (TII), we selected the first four tasks—NCR, HTU, MTA, and BRI—for exploration, leaving TCU and TII for rigorous out-of-domain testing. We follow the same procedure indicated in \Cref{subsubsec:toolbench_query_instances} to generate 400 query instances, all formatted consistently.
\section{Experimental Setup} \label{experiments-all}
With the integration of environments into \method~ detailed in \Cref{sesc:applications}, we conducted experiments to demonstrate our framework's effectiveness on ToolBench and CRMArena benchmarks.

\label{sec:exps}
\subsection{Evaluation datasets and metrics}
\paragraph{ToolBench Evaluation.} To comprehensively assess ToolBench's performance, we employed three distinct test sets with 600 instances tailored to examine different scenarios. The first test set, \emph{Unseen Instruction}, or \texttt{G1\_inst}, is designed to measure how well the model performs when presented with new instructions for potentially familiar tools. It is important to highlight that due to the small number of tasks and tools selected for exploratory purposes, as detailed in \Cref{subsubsec:toolbench_query_instances}, the likelihood of encountering similar instructions or tools in this set is exceedingly low. The second test set, \emph{Unseen Tools}, or \texttt{G1\_tool}, evaluates the model's ability to manage tools it has never encountered before. Finally, the \emph{Unseen Tools \& Unseen Categories}, or \texttt{G1\_cat}, test set presents the most challenging scenario by testing the model on tasks from entirely new categories, requiring the use of unfamiliar tools.

\paragraph{CRMArena Evaluation.} In evaluating CRMArena, we utilized public test sets corresponding to the six tasks detailed in \Cref{subsec:crmarena_tasks}. Each task contains 130 entirely new instances to the exploration data. The NCR, HTU, MTA, and BRI tasks, while sharing similar scenarios or tools encountered during exploration, are distinguished by their novel use cases. In contrast, the TCU and TII tasks are entirely distinct from any previously explored domain, ensuring these tests are entirely out-of-domain.

\subsection{Agent LLM}
\paragraph{ToolBench.} In our study, we focused on the two leading models in the ToolBench benchmark: \texttt{gpt-4o} ~\cite{achiam2023gpt} and \texttt{xlam-8x7b} ~\cite{zhang2024xlam}. For a more comprehensive analysis, we also included their more compact counterparts, \texttt{gpt-4o-mini} and \texttt{xlam-7b-r}, in our exploration and fine-tuning processes. To assess a broader range of performance capabilities, we additionally experimented with the lower-performing model, \texttt{mixtral-8x7b-inst} ~\cite{jiang2024mixtral}.

\paragraph{CRMArena.} In the CRMArena environment, which demands advanced planning and the complex use of tools, we chose the best-performing model, \texttt{gpt-4o}, along with its compact version, \texttt{gpt-4o-mini}, as our baseline LLM Agents for exploration and fine-tuning. Given these requirements, we observed that lower-performing models such as \texttt{mixtral-8x7b-inst} struggled to produce effective trajectories. Thus, constrained by time and resources, we opted not to experiment with models failing to meet the environment's high-performance demands.

\subsection{Training Datasets}
We incorporated the Query Instances from Section \Cref{sesc:applications} for exploration. We let the Agent LLM to continuously explore the given tasks with temperature 1.0, and collected all trajectories that passed our evaluators, until it reached 500 trajectories for training. The filtered trajectories then directly being used to finetune the same base model.

\subsection{Comparison with other work}
To highlight the effectiveness of \method~, we conducted a comparative analysis against the ToolLLM approach~\cite{qin2023toolllm}. ToolLLM pioneered the use of large language models (LLMs) to automatically generate and evaluate tasks for agentic training data. For a fair comparison, we reproduced two versions of their methodology: \texttt{ToolLLM\_Full}, which faithfully follows the original process using GPT-4o and GPT-4o-mini to both generate and evaluate tasks, and \texttt{ToolLLM\_Partial}, which combines our high-quality task generation process with ToolLLM’s LLM-based evaluation procedure. We selected CRMArena as the testing environment because this benchmark presents highly complex and challenging tasks, making it ideal for a robust evaluation of both methods. This setup allows us to clearly assess the impact of our data generation and evaluation improvements.

\section{Results and Discussions}
\begin{table*}
\small
  \centering
  \begin{tabularx}{\linewidth}{lXXXX}
    \hline
{Model Name} & {ALL} & {\texttt{G1\_inst}} & {\texttt{G1\_cat}} & {\texttt{G1\_tool}} \\ \hline
\rowcolor{cyan!30} \texttt{gpt-4o-ls (ours)} & 0.515 & 0.465 & 0.555 & 0.525 \\ 
\texttt{gpt-4o} & 0.474 & 0.437 & 0.519 & 0.466 \\ \hline

\rowcolor{cyan!30} \texttt{xlam-8x7b-ls (ours)} & 0.498 & 0.440 & 0.530 & 0.525 \\ 
\texttt{xlam-8x7b-r} & 0.435 & 0.415 & 0.490 & 0.400 \\ \hline

\rowcolor{cyan!30} \texttt{gpt-4o-mini-ls (ours)} & 0.497 & 0.475 & 0.540 & 0.475 \\ 
\texttt{gpt-4o-mini} & 0.452 & 0.415 & 0.505 & 0.435 \\ \hline

\rowcolor{cyan!30} \texttt{xlam-7b-ls (ours)} & 0.413 & 0.400 & 0.460 & 0.380 \\ 
\texttt{xlam-7b-r} & 0.392 & 0.355 & 0.425 & 0.395 \\ \hline

\rowcolor{cyan!30} \texttt{mixtral-8x7b-ls (ours)} & 0.310 & 0.280 & 0.385 & 0.265 \\ 
\texttt{mixtral-8x7b-inst} & 0.117 & 0.085 & 0.160 & 0.105 \\ \hline
  \end{tabularx}
  \caption{
  \small
Pass Rate (\%) on three distinct ToolBench test sets. "ALL" denotes the average performance across all test sets. Models are categorized into baseline versions and their fine-tuned counterparts, with models trained using self-generated data through \method~ highlighted in \textcolor{cyan}{cyan}.}
  \label{tab:tooleval_result}
\end{table*}


\subsection{ToolBench}
The evaluation of the ToolBench datasets, as presented in Table \ref{tab:tooleval_result}, demonstrates substantial performance improvements in our fine-tuned models relative to their baselines. The \texttt{gpt-4o-ls} model shows a marked improvement, increasing its performance from 47.4\% to 51.5\% compared to the baseline \texttt{gpt-4o}. Similarly, the \texttt{xlam-8x7b-ls} model exhibits a notable enhancement, rising from 43.5\% to 49.8\% over its baseline, \texttt{xlam-8x7b-r}. Among compact models, \texttt{gpt-4o-mini-ls} and \texttt{xlam-7b-ls} achieved performance gains of 4.5\% and 2.1\%, respectively, when compared to their baselines. These improvements are particularly impressive because these models already rank among the best in the benchmark. Notably, low-performing models benefited even more from our approach. The \texttt{mixtral-8x7b-inst} model, initially achieving an 11.7\% pass rate, improved significantly to 31.0\% after fine-tuning to the \texttt{mixtral-8x7b-ls} version. This demonstrates the effectiveness of self-generated data in enhancing model performance.

When analyzing the performance across different test sets, we observed substantial gains in out-of-domain tasks. This is evident in the improvements on the G1\_cat (Unseen Tools in Unseen Category) and G1\_tool (Unseen Tool) datasets. For instance, top-performing model \texttt{xlam-8x7b-ls} gained 4\% on G1\_cat and 12.5\% on G1\_tool compared to its baseline. Similarly, the lower-performing \texttt{mixtral-8x7b-ls} model recorded more than double gains on both G1\_cat and G1\_tool over its baseline. These results highlight the efficacy of our framework in producing high-quality data for enhancing agentic learning.

\subsection{CRMArena}
\begin{table*}
  \small
  \centering
  \begin{tabularx}{\linewidth}{lXXXXXXX}
    \hline
{Model Name} & {ALL} & {NCR} & {HTU} & {MTI} & {BRI} & {TCU} & {TII} \\ \hline
\rowcolor{cyan!30} \texttt{gpt-4o-ls (ours)} & 0.864 & 0.677 & 0.808 & 0.985 & 0.869 & 0.862 & 0.985 \\
\texttt{gpt-4o-ToolLLM\_partial} & 0.768 & 0.285 & 0.785 & 0.969 & 0.762 & 0.854 & 0.954 \\
\texttt{gpt-4o-ToolLLM\_full} & 0.546 & 0.120 & 0.269 & 0.815 & 0.446 & 0.685 & 0.938 \\
\texttt{(base) gpt-4o} & 0.623 & 0.600 & 0.477 & 0.277 & 0.592 & 0.815 & 0.977 \\ \hline
\rowcolor{cyan!30} \texttt{gpt-4o-mini-ls (ours)} & 0.678 & 0.262 & 0.715 & 0.762 & 0.485 & 0.877 & 0.969 \\
\texttt{gpt-4o-mini-ToolLLM\_partial} & 0.455 & 0.038 & 0.277 & 0.454 & 0.431 & 0.569 & 0.962 \\
\texttt{gpt-4o-mini-ToolLLM\_full} & 0.232 & 0.023 & 0.246 & 0.446 & 0.192 & 0.346 & 0.138 \\
\texttt{(base) gpt-4o-mini} & 0.185 & 0.080 & 0.108 & 0.000 & 0.215 & 0.108 & 0.600 \\ \hline
  \end{tabularx}
  \caption{\small
Pass Rate (\%) on six distinct CRMArena test sets. "ALL" denotes the average performance across all test sets. Models are shown as baseline versions and as those fine-tuned using either the ToolLLM or our own methodology. For ToolLLM models, \texttt{ToolLLM\_full} follows the original approach of leveraging LLMs (GPT-4o and GPT-4o-mini) for both task generation and evaluation, while \texttt{ToolLLM\_partial} uses our high-quality tasks with LLM-based evaluation. Models trained with self-generated data from \method~ are highlighted in \textcolor{cyan}{cyan}.}
  \label{tab:crmarena}
\end{table*}

For CRMArena test sets, which require advanced problem-solving abilities in complex, realistic CRM environments, our fine-tuned model, \texttt{gpt-4o-ls}, demonstrates a marked improvement. It achieves an overall pass rate of 86.4\%, representing a significant increase of 24.1\% over its baseline version, \texttt{gpt-4o}, which scores 62.3\%. This enhancement is even more pronounced in the weaker model, where our framework boosts \texttt{gpt-4o-mini}'s performance nearly fourfold, elevating the accuracy from 18.5\% to 67.8\%. Such a transformation turns a previously inadequate model into one that is highly effective for these tasks.

When compared to the ToolLLM methodologies, our approach consistently outperforms both \texttt{ToolLLM\_Full} and \texttt{ToolLLM\_Partial} across all test sets. For example, our model \texttt{gpt-4o-ls} achieves an average pass rate that is over 31\% higher than \texttt{gpt-4o-ToolLLM\_full} (54.6\%) and nearly 10\% higher than \texttt{gpt-4o-ToolLLM\_partial} (76.8\%). Importantly, for the mini-model variants, the gains are even more substantial: \texttt{gpt-4o-mini-ls} surpasses \texttt{gpt-4o-mini-ToolLLM\_full} (23.2\%) and \texttt{gpt-4o-mini-ToolLLM\_partial} (45.5\%), confirming the robustness and effectiveness of our data generation framework, especially in low-resource settings.

Crucially, these performance improvements extend beyond just in-domain tasks to also significantly impact out-of-domain tasks. \texttt{gpt-4o-mini-ls} shows notable gains, achieving a 76.9\% increase on the TCU task and a 36.9\% increase on the TII task. Compared to the relatively modest gains seen with ToolLLM variants, these results illustrate a substantial enhancement in the model's understanding and generalization of CRM tools and associated tasks, both within and beyond the immediate training domain. Overall, our method clearly leads to superior task competence and better generalization than existing approaches.





\subsection{Ablation Studies}
\subsubsection{Tools usage errors reduction}
Besides discussing the effectiveness of \method~ in enhancing agents' overall problem-solving capabilities, we also examined its utility in improving agents' tool usage. Our study involved models \texttt{xlam-8x7b-ls} and \texttt{xlam-7b-ls}, compared against their baseline counterparts \texttt{xlam-8x7b-r} and \texttt{xlam-7b-r}. We randomly selected 200 states, each with preceding steps generated by one of these four models, tasked with solving tasks from the ToolBench test sets. Each model's subsequent actions were assessed for errors using our Action Handler.

Our analysis focused on three layers of errors: 1) structural errors, where actions are unparseable; 2) toolname errors, where actions are parseable but the tool names are hallucinated; and 3) arguments errors, where both parsing and tool names are correct, but arguments are misused.

\begin{table}
{\fontsize{7.5}{9.5}\selectfont
  \centering
  \begin{tabularx}{\linewidth}{lXXXX}
    \hline
{Model Name} & {ALL} & {Structure} & {Toolname} & {Arguments} \\ \hline
\rowcolor{cyan!30} \texttt{xlam-8x7b-ls} & 16.67 & 25 & 1 & 24 \\
\texttt{xlam-8x7b-r} & 25.33 & 30 & 6 & 40 \\ \hline
\rowcolor{cyan!30} \texttt{xlam-7b-ls} & 13.67 & 17 & 12 & 12 \\
\texttt{xlam-7b-r} & 35.00 & 42 & 8 & 55 \\
  \end{tabularx}}
  \caption{
  \small
Number of errors (lower is better) for actions generated from 200 sampled states across ToolBench test sets, comparing different models. Error types include structural errors (Structure), hallucinated tool calls (Toolname), and incorrect tool argument usage (Arguments). "ALL" indicates overall average number of errors. Models fine-tuned via \method~ are highlighted in \textcolor{cyan}{cyan}.
}
  \label{tab:error_analysis}
\end{table}

The results in Table \ref{tab:error_analysis} illustrates that \method~ substantially reduces errors, nearly halving them on average. For the compact model \texttt{xlam-7b-r}, substantial reductions were observed particularly in structural errors and incorrect tool argument errors. Though there's a slight increase in tool hallucination, it can be attributed to the baseline model primarily committing structural errors, resulting in previously obscured toolname errors being revealed. Even among top-performing models, notable reductions in argument errors were observed. This suggests that \method~ effectively enhances agents' tool usage capabilities.

\subsubsection{Impact of Core Monitoring Components}
We conducted an ablation study to evaluate the impact of core monitoring components in our \method~ framework. Specifically, we sought to understand the contributions of \textbf{Action handler}, which monitors the LLM Agent's ability to perform correct actions through tool usage, and \textbf{Trajectory filtering}, which examines the agent's overall task-solving ability by comparing its trajectory against a ground-truth solution.

We used the \texttt{mixtral-8x7b-inst} as our baseline LLM Agent. We utilized \method~ with the LLM Agent as it interacted with tasks from ToolBench, as specified in \Cref{subsubsec:toolbench_query_instances}, similar to the primary experiment. However, we made a critical modification by disabling the Action Handler when collecting training trajectories. That means, the Agent's actions, particularly those involving tool calls, were not monitored. We then fine-tuned the baseline model with the self-generated data and evaluated on the three ToolBench test sets used in the main experiment, enabling us to analyze the impact of \textbf{"Not monitor Action"}.

Similarly, we started with the same baseline LLM Agent and pass through \method~ for generating data while disabling ground-truth comparison within Trajectory filtering. In this approach, there was no quality control for trajectories misaligned with the task requirements. Subsequently, we fine-tuned the baseline model and performed evaluations on the same three ToolBench test sets. This allowed us to assess the impact of \textbf{"Not monitor Trajectory"}.

\begin{figure}
  \begin{center}
\includegraphics[width=\linewidth]{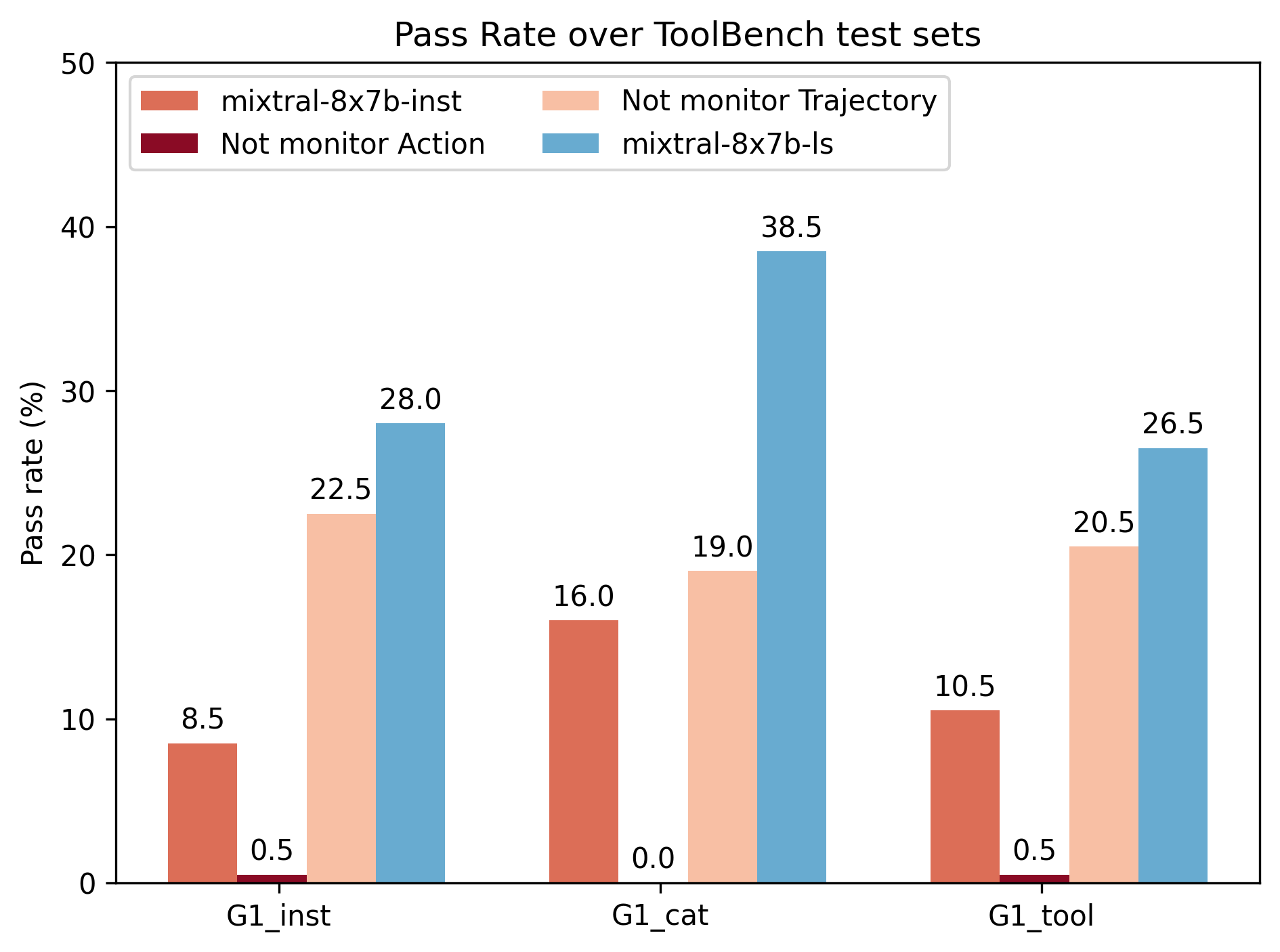}
  \end{center}
    \caption{\small Ablation study on monitoring components across three ToolBench test sets. \texttt{mixtral-8x7b-inst} shows baseline model performance. "Not monitor Action" indicates performance when fine-tuning with a self-exploration dataset via \method~ without action monitoring, while "Not monitor Trajectory" shows results without trajectory monitoring. \texttt{mixtral-8x7b-ls} demonstrates performance with both action and trajectory monitoring enabled.}
  \label{fig:ablation_components}
  \vspace{-3mm}
\end{figure}

Our ablation study, depicted in Figure \ref{fig:ablation_components}, underscores the essential role of both action-level and trajectory-level monitoring in our model's performance. The removal of action-level monitoring ("Not monitor Action" in Figure \ref{fig:ablation_components}) results in a significant decline in pass rates across all test sets. This suggests that without proper regulation of tool usage, errors accumulate, which the model is unable to rectify autonomously. Similarly, the absence of trajectory-level monitoring ("Not monitor trajectory") hinders the overall effectiveness of the model by failing to ensure alignment with task requirements. This shortcoming is particularly evident in the out-of-domain G1\_cat setting, where the pass rate drastically decreases from 38.5\% to 19.0\%. These findings underscores the importance of monitoring the actions and outcomes of LLM Agents in scenarios where we aim to apply self-improvement to agentic models, thus, clearly demonstrate the importance of incorporating all monitoring components in our \method~.
\section{Conclusion}

In this paper, we presented \method~, a comprehensive framework designed to advance the development of Large Action Models (LAMs) by enabling self-learning through online exploration and automated prorgammatic feedback. Our system effectively addresses the limitations of traditional supervised learning and manual data curation, offering a scalable solution that enhances both agentic performance and training efficiency. \method~ provides real-time interactions, multi-turn task processing, and high-quality feedback, contributing to significant improvements in model training performance across various benchmarks, such as ToolBench and CRMArena, where models trained with self-generated data via \method~ gained a significant improvements and potentially outperformed other leading models. Our framework accelerates the learning and adaptation process of LAMs with minimal human intervention, demonstrating its potential as a pivotal tool for future research and development in AI agents.
\section{Limitations}
\method~ still has some limitations, its current implementation focuses on predefined tasks and tools, which may limit its adaptability in more dynamic or unstructured environments. 
In future work, we aim to expand the framework's generalization capabilities by incorporating a wider range of tasks and tool integrations, as well as exploring methods for better handling ambiguous or incomplete task specifications. Furthermore, we plan to investigate the scalability of the system in environments with more complex action spaces and interdependencies, pushing the boundaries of autonomous agent learning.

\bibliography{acl_latex}

\clearpage

\appendix

\section{Appendix}
\label{sec:appendix}
\subsection{Examples of Query Instance}

\label{subsec:example_query_instance}
\paragraph{Query template}
Here, we are providing an example for query templates. As we can see from Figure \ref{fig:example_query_template}, a query template contains natural text portions and placeholders to-be-filled.

\begin{figure}[h]
\begin{lstlisting}[language=json,basicstyle=\scriptsize\ttfamily, backgroundcolor=\color{lightgold!50}]
"query_template": "I've been looking up {movie_detail} about the movie {movie_name}. Fun fact: the set of {movie_name} was built inside a massive warehouse to create a surreal atmosphere!"
\end{lstlisting}
\caption{Example of Query Templates}
\label{fig:example_query_template}
\end{figure}

\paragraph{Placeholders description}
We then provide example for query template's placeholders' descriptions (Figure \ref{fig:example_query_params_desc}).
\begin{figure}[h]
\begin{lstlisting}[language=json,basicstyle=\scriptsize\ttfamily, backgroundcolor=\color{lightgold!50}]
"placeholders_metadata": {
    "movie_name": {
        "type": str,
        "description": "The name of the movie to search for.",
    },
    "movie_detail":  {
        "type": str,
        "description": "The detail of the movie to search for.",
    }
}
\end{lstlisting}
\caption{Example of Placeholders description}
\label{fig:example_query_params_desc}
\end{figure}

\paragraph{Generated Placeholders}
Given the query template and its placeholders' descriptions, we can use Large Language Models (LLMs) for generating dynamic values for the placeholders. An example of a generated placeholder is shown in Figure 
\ref{fig:example_query_params_gen}.

\begin{figure}[h]
\begin{lstlisting}[language=json,basicstyle=\scriptsize\ttfamily, backgroundcolor=\color{lightgold!50}]
"placeholders": {
    "movie_name": "The Dark Knight",
    "movie_detail": "genres"
}
\end{lstlisting}
\caption{Example of generated Placeholders}
\label{fig:example_query_params_gen}
\end{figure}

\paragraph{Filled-in query} With Query Template (\ref{fig:example_query_template}), and Generated Placeholders (\ref{fig:example_query_params_gen}), we can fill in the value of the placeholder into the query template to create Filled-in query. An example is showned at \ref{fig:example_filled_in_query}.

\begin{figure}[h]
\begin{lstlisting}[language=json,basicstyle=\scriptsize\ttfamily, backgroundcolor=\color{lightgold!50}]
"filled_in_query": "I've been looking up genres about the movie The Dark Knight. Fun fact: the set of The Dark Knight was built inside a massive warehouse to create a surreal atmosphere!"
\end{lstlisting}
\caption{Example of Filled-in Query with Query template \ref{fig:example_query_template} and generated Placeholders \ref{fig:example_query_params_gen}. Note that after this, the query can be further paraphrased with LLM for diversity purpose.}
\label{fig:example_filled_in_query}
\end{figure}

\paragraph{Answer computation} We also give an example of how we can compute answer for the generated query \ref{fig:example_filled_in_query}. As illustrated in Figure \ref{fig:example_answer_compute}, we pre-define a solution path for any task that can be formed with the query template. This solution path will then being used by any generated task with the query template \ref{fig:example_query_template} for ground-truth answer computation.

\begin{figure}[h]
\begin{lstlisting}[language=json,basicstyle=\scriptsize\ttfamily, backgroundcolor=\color{lightgold!50}]
"solution_paths": [
    {
        "tool_call": "get_search_movie_for_movie_tools",
        "arguments": {
          "movie_name": null
        }
    },
    {
        "tool_call": "get_movie_details_for_movie_tools",
        "arguments": {
          "id": null
        }
    }
]
\end{lstlisting}
\caption{Example of a solution path for the task \ref{fig:example_filled_in_query}. The arguments would be searched among 1) placeholder values and 2) objects generated during execution. In this example, \texttt{movie\_name} can be extracted directly from the placeholder value (The Dark Night), from \ref{fig:example_query_params_gen}, while \texttt{id} is a new field can be retrieved from the execution response of \texttt{get\_search\_movie}.
}
\label{fig:example_answer_compute}
\end{figure}

\paragraph{Available tools} We then provide an example of a set of available tools provided to LLM Agents for exploration. As we can see from \ref{fig:example_available_tools}, the tools set does not include \texttt{get\_search\_movie\_for\_movie\_tools}, but instead include an alternative version \texttt{search\_movie\_for\_imdb}, which does the similar objective with \texttt{get\_search\_movie}, but it is a different tool with different way to use. In addition, there are extra tools provided here to, where the LLM Agent is expected to decide what is the right tool to use at a given time step.

\begin{figure}[h]
\begin{lstlisting}[language=json,basicstyle=\scriptsize\ttfamily, backgroundcolor=\color{lightgold!50}]
"task_available_tools": [
  "get_movie_details_for_movie_tools",
  "search_movie_for_imdb",
  "get_movie_production_companies_for_movie_tools",
  "get_current_temp_for_weather_tools",
  "Finish"
]
\end{lstlisting}
\caption{\small Example set of available tools to be provided for LLM Agent for exploration. Note that this tools set does not include \texttt{get\_search\_movie\_for\_movie\_tools}, but instead include an alternative version \texttt{search\_movie\_for\_imdb}, which does the similar objective with \texttt{get\_search\_movie}, but it is a different tool with different way to use. In addition, there are extra tools provided here to, where the LLM Agent is expected to decide what is the right tool to use at a given time step.}
\label{fig:example_available_tools}
\end{figure}

\subsection{Example of LLM Agent system prompt}
We also include an example of a system prompt for LLM Agent that we derived from ToolBench's ~\cite{qin2023toolllm} and xLAM's ~\cite{zhang2024xlam} system prompt, as shown in \ref{fig:example_agent_prompt}.

\subsection{Clean-up Toolbench tools}
\label{subsec:cleanup_toolbench_tools}

\begin{figure}[h]
\centering
\includegraphics[width=0.45\textwidth]{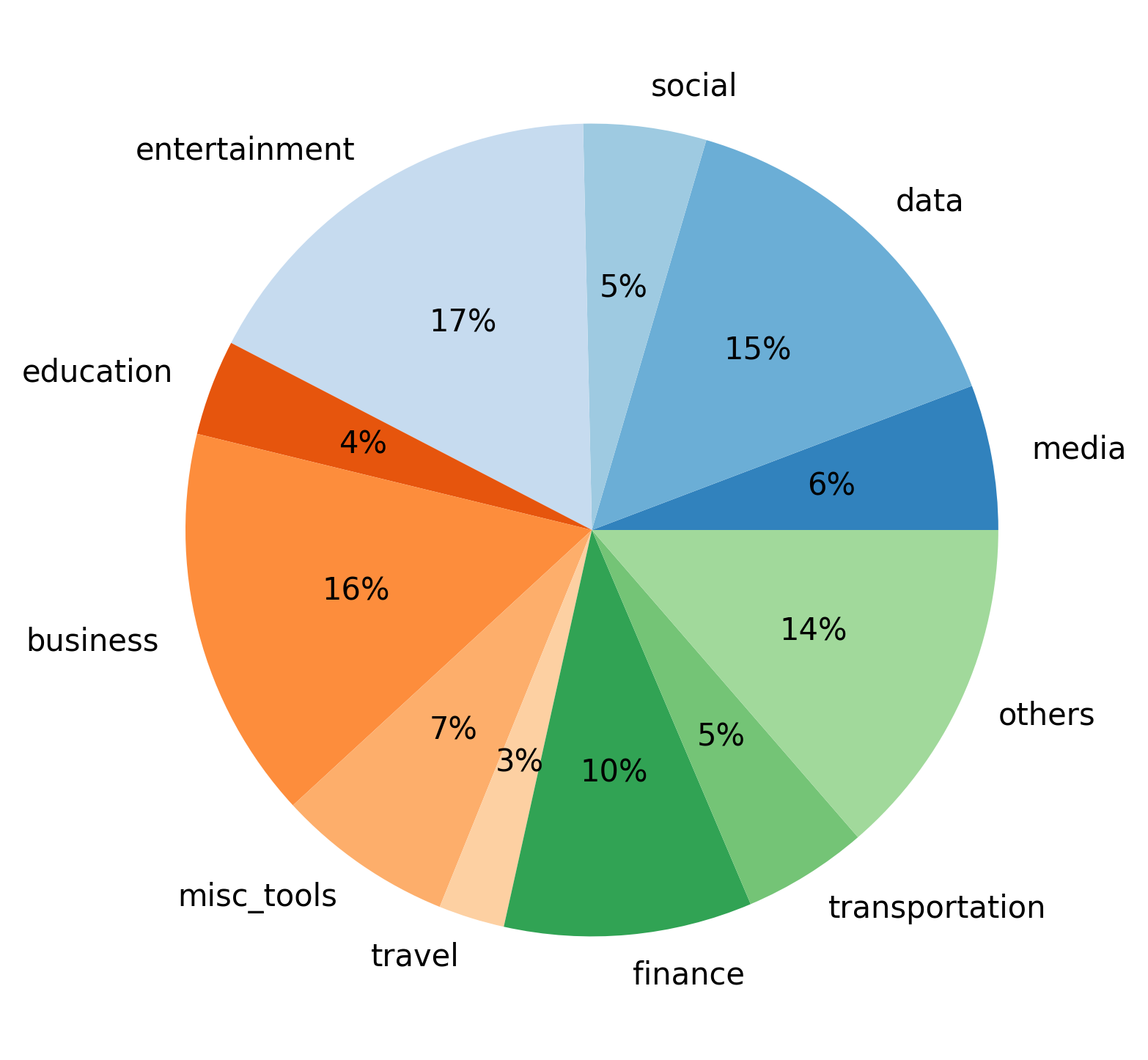}
\caption{Tools distribution for ToolBench environment. }
\label{fig:tools_distribution}
\end{figure}

We leveraged ToolLLM’s extensive work under Apache-2.0 license, which collated over 16,464 REST APIs across 49 categories from RapidAPI Hub. Despite the breadth, the quality and documentation of these tools were inconsistent due to their mass collection approach, leading to many non-functional or poorly documented entries. To rectify this, critical elements such as tool names, parameters, execution code, and related metadata were extracted. Large language models (LLMs) played a crucial role in refining the tool descriptions and docstrings to ensure clarity and coherence. This process involved integrating necessary Python code components, conducting validations for code executability, and leveraging LLMs to assess quality. Moreover, rule-based techniques and LLM prompting were used to eliminate duplicate or similar tools, enhancing the collection’s integrity. In Figure \ref{fig:tools_distribution}, we display the distribution of our tools collection for ToolBench after cleaning up and rewriting documentations.

Note that for now, we are only utilizing a small subset of this collection for exploration. In the future, we are going to scale up to try to use all of the processed tools.

\begin{figure*}[ht]
\begin{lstlisting}[language=json,basicstyle=\scriptsize\ttfamily, backgroundcolor=\color{lightgold!50}]
[BEGIN OF TASK INSTRUCTION]
You are an expert in agentic task. You will be given a task, and you can use many tools sequentially to solve the task. At each time step, you will call exactly 1 tool, and based on the environment feedback, you will be able to decide your next step. Keep repeating this action until you gather enough information to solve the task. By that time, call the special function "Finish" given to use to return the final answer in the exact format.

Remember:
1. MOST IMPORTANT, in your response of the "Finish" step, you MUST strictly follow the response format of what to be written inside "final_answer".
2. The state change is irreversible, you can't go back to one of the former state.
3. All the thought is short, at most in 5 sentences.
4. Your action must be calling one of the given tools (functions).
5. Your action input must be in json format, where action inputs must be realistic and from the user. Never generate any action input by yourself or copy the input description. Do not add unrelated parameters if not needed. Do not add optional parameters when it is not required or when these information is not needed.
6. You can do more then one trys, so if your plan is to continusly try some conditions, you can do one of the conditions per try.

Task description:
You should use functions to help handle the real time user querys. Remember:
1. ALWAYS call "Finish" function at the end of the task. And the final answer should contain enough information to show to the user.
[END OF TASK INSTRUCTION]
\end{lstlisting}
\caption{\small Example of system prompt for LLM Agent}

\label{fig:example_agent_prompt}
\end{figure*}

\section{Licenses}
Here, we discussed about the licenses of the artifacts we used in our work.

For tools (code logic), we used tools from ToolBench and CRMArena. ToolBench is under Apache-2.0 License, and CRMArena is under Creative Commons Attribution 4.0 License (CC BY).

For evaluation datasets, we also used the datasets from ToolBench and CRMArena, in which licenses are mentioned above.

For models we used for exploration (data generation), finetuning, and evaluating, we used \texttt{gpt-4o}, \texttt{gpt-4o-mini}, \texttt{xlam-8x7b-r}, \texttt{xlam-7b-r}, \texttt{mixtral-8x7b-inst}. Here:
\begin{itemize}
    \item \texttt{gpt-4o} and \texttt{gpt-4o-mini} ~\cite{achiam2023gpt} is under MIT License.
    \item \texttt{xlam-8x7b-r} and \texttt{xlam-7b-r} ~\cite{zhang2024xlam} is under Apache 2.0 License. \item \texttt{mixtral-8x7b-inst} ~\cite{jiang2024mixtral} is under Apache 2.0 License.
\end{itemize}. 

All of the licenses above enable us to perform research experiments.

\section{Experimental Detail}
Our data generation and trainings of \texttt{mixtral-8x7b-inst} (56B parameters), \texttt{xlam-8x7b-r} (56B parameters), and \texttt{xlam-7b-r} (7B parameters) are performed with 4*H100s machines. Each of the exploration iteration to generate data is limited to 8 hours, and corresponding training time is limited by 4 hours. The hyperparameters of training all instances are at $5e-6$ for 3 epochs.

For the data generations and trainings on \texttt{gpt-4o} and \texttt{gpt-4o-mini}, we used OpenAI's endpoint with the same time limit. For trainings, we used default hyperparameters and number of epochs suggested by OpenAI, which is at between 1 and 2 for LR multiplier and for 3 epochs.

For evaluations, we configured the generative temperature to be 0.0. This allows us to have deterministic results for the presented ones in Section \ref{experiments-all}.

\section{Others}
When constructing this paper, we used \texttt{gpt-4o} ~\cite{achiam2023gpt} for several paraphrasing.

\end{document}